\documentclass{article}
\usepackage{ijcai26}

\usepackage{times}
\usepackage{soul}
\usepackage{url}
\usepackage[hidelinks]{hyperref}
\usepackage[utf8]{inputenc}
\usepackage[small]{caption}
\usepackage{graphicx}
\usepackage{amsmath}
\usepackage{amsthm}
\usepackage{booktabs}
\usepackage{algorithm}
\usepackage{algorithmic}
\usepackage[switch]{lineno}
\usepackage{array}
\usepackage{multirow}
\usepackage{booktabs}
\usepackage{multirow}
\usepackage{array}
\usepackage{xcolor}   
\usepackage{xspace}   
\usepackage{amssymb}

\title{DynaTrust: Defending Multi-Agent Systems Against Sleeper Agents via\\ Dynamic Trust Graphs}

\author{
Yu Li$^1$
\and
Qiang Hu$^1$\and
Yao Zhang$^1$\and
Lili Quan$^2$\and
Jiongchi Yu$^2$\And
Junjie Wang$^1$\\
\affiliations
$^1$Tianjin University\\
$^2$Singapore Management University\\
\{liyu2025, qianghu\}@tju.edu.cn
}

\begin{document}

\maketitle

\begin{abstract}
Large Language Model-based Multi-Agent Systems (MAS) have demonstrated remarkable collaborative reasoning capabilities but introduce new attack surfaces, such as the sleeper agent, which behave benignly during routine operation and gradually accumulate trust, only revealing malicious behaviors when specific conditions or triggers are met.
Existing defense works primarily focus on static graph optimization or hierarchical data management, often failing to adapt to evolving adversarial strategies or suffering from high false-positive rates (FPR) due to rigid blocking policies. To address this, we propose \textit{DynaTrust}, a novel defense method against sleeper agents. \textit{DynaTrust} models MAS as a dynamic trust graph~(DTG), and treats trust as a continuous, evolving process rather than a static attribute. It dynamically updates the trust of each agent based on its historical behaviors and the confidence of selected expert agents. Instead of simply blocking, \textit{DynaTrust} autonomously restructures the graph to isolate compromised agents and restore task connectivity to ensure the usability of MAS. To assess the effectiveness of DynaTrust, we evaluate it on mixed benchmarks derived from AdvBench and HumanEval. The results demonstrate that DynaTrust outperforms the state-of-the-art method AgentShield by increasing the defense success rate by 41.7\%, achieving rates exceeding 86\% under adversarial conditions. Furthermore, it effectively balances security with utility by significantly reducing FPR, ensuring uninterrupted system operations through graph adaptation. 

\end{abstract}

\section{Introduction}

The rapid evolution of Large Language Models (LLMs) has transformed intelligent systems from standalone conversational agents to autonomous Multi-Agent Systems (MAS), where multiple agents collaborate to perform complex tasks in software development, decision making, and simulation~\cite{Hong2023MetaGPTMP,Qian2023ChatDevCA,Wu2023AutoGenEN,Li2023CAMELCA}. By coordinating reasoning and execution across agents with differentiated roles and expertise, MAS can achieve problem-solving capabilities that surpass those of individual models~\cite{agentsdebate,GenerativeAgents,multiagent}.  

However, this collaborative structure also introduces a critical security risk similar to insider attack: \emph{sleeper agents}~\cite{sleeper,Sleeperagent}, which behave benignly during routine operation and gradually accumulate trust, only revealing malicious behaviors when specific conditions or triggers are met. This risk is amplified by inherent vulnerabilities in LLM-based agents, which are prone to hallucinations and susceptible to social engineering and strategic manipulation~\cite{Zhang2024PsySafeAC,Wei2023JailbrokenHD}. 
In a MAS, a single compromised agent can exploit the network’s semantic and structural connections to spread malicious instructions, coordinate jailbreak strategies, and bypass safety filters through seemingly legitimate interactions~\cite{Jailbreaks,llmjailbreaking,lee2025prompt,prompt}. Such insider-style attacks are harmful but hard to detect as they originate from within the system and remain hidden until substantial damage has already occurred.

Although important, no specific defense methods have been designed for sleeper agents. Existing defense mainly focuses on defending against static malicious injection and jailbreak attacks. Directly employing these methods for sleeper agents faces several challenges.
(1) Existing defense methods largely rely on fixed security policies~\cite{yu-etal-2025-netsafe,mao2025agentsafesafeguardinglargelanguage}, making zero-trust decisions based on short-term observations or isolated anomalies. Such designs fail to account for the inherent stochasticity of LLM behaviors, including occasional hallucinations, resulting in high false positive rates. Moreover, once an agent is flagged, these methods typically enforce permanent blocking without mechanisms for trust recovery, which degrades system availability and disrupts normal MAS operation. (2) Most existing approaches adopt majority voting or consensus mechanisms that treat all agents as equally reliable~\cite{agentsdebate,survey}. 
This overlooks variations in agent expertise and historical behavior, and fails to quantify decision uncertainty in ambiguous contexts.

\begin{figure}[t]
    \centering
    \includegraphics[width=\linewidth]{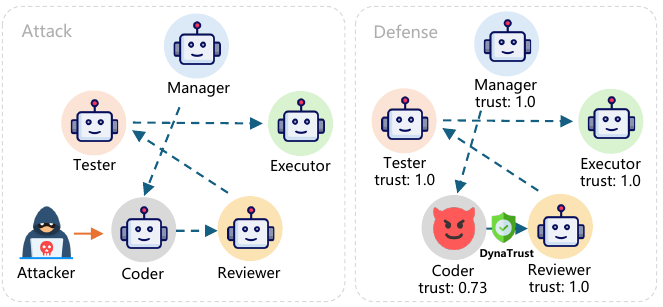} 
        \caption{The attack and DynaTrust defense models of MAS.}
    \label{fig:attack&defense}
\end{figure}

To address these challenges, we propose \textit{DynaTrust}, a Dynamic Trust-Driven Consensus Defense framework that protects MAS against sleeper agents through dynamic trust evolution, trust-confidence weighted consensus, and adaptive graph recovery. DynaTrust models MAS as a dynamic trust graph (DTG) and treats trust as a continuous, evolving process rather than a static attribute as shown in Figure \ref{fig:attack&defense}. It consists of three core mechanisms. (1) It models each agent’s reliability as a probabilistic trust state, integrating historical interactions with a Bayesian penalty scheme to smooth short-term fluctuations while rapidly penalizing persistent malicious behavior. (2) It then dynamically selects a targeted jury of semantic expert agents and weights their votes by both trust history and real-time confidence, enabling precise decision making even under uncertainty. (3) Finally, agents whose trust falls below a critical threshold are automatically isolated, and instantiate replica agents are activated to restore system connectivity and preserve operation without human intervention. 

To evaluate the usefulness of \textit{DynaTrust}, we integrates it into 4 MAS and perform it on mixed benchmarks derived from AdvBench and HumanEval. Compared to the state-of-the-art defense method, AgentShield, \textit{DynaTrust} demonstrates superior defense success rate with an average of 41.7\% improvement. Ablation studies showcase that each mechanism positively contributes to \textit{DynaTrust}. 

Our contributions are summarized as follows:
\begin{itemize}

\item We model multi-agent systems as dynamic trust graphs, enabling the modeling of long-horizon insider risks through trust evolution.

\item We propose \textit{DynaTrust}, a novel defense method that dynamically updates agent trust, performs trust-aware consensus, and restores system connectivity to defend against sleeper-agent attacks in MAS.

\item Extensive evaluations on AdvBench and HumanEval demonstrate that DynaTrust achieves strong DSR while maintaining high system availability.
\end{itemize}

\section{Related Work}
\subsection{Large Language Model Safety}
\textbf{Adversarial Attacks and Jailbreaking.} Early attacks relied on hand-crafted ``jailbreak prompts'' (like {\em Do Anything Now}) to bypass safety filters~\cite{Shen2023DAN}. Recently, automated attack frameworks have become more sophisticated. ~\cite{advbench} proposed a gradient-based method to automatically generate universal attack suffixes that work across models like Llama-2 and GPT-4. Similarly, ~\cite{JailbreakingBB} and~\cite{MASTERKEY} used one LLM to ``red-team'' another, automatically finding loopholes to generate harmful content. Furthermore, ~\cite{FinetuningAL} showed that fine-tuning an LLM on just a few malicious examples can completely break its safety alignment. These attacks on single models serve as the fundamental building blocks for threats in multi-agent systems.

\textbf{Single-Model Defenses.} To counter these threats, the community has developed various defenses. The most common approach is {\em Reinforcement Learning from Human Feedback}    (RLHF)~\cite{TrainingLM}, which trains models to refuse harmful instructions. At inference time, methods like {\em Perplexity-based Detection}~\cite{DetectingLM} and {\em SmoothLLM}~\cite{smoothllm} try to identify attacks by checking for unusual input patterns or adding random noise. ~\cite{DefendingCA} encourage models to ``reflect'' on safety guidelines before answering. However, these defenses assume a single-turn or independent interaction. As noted by ~\cite{yu-etal-2025-netsafe}, they often fail in multi-agent settings where malicious intent is split across multiple agents or hidden within a long collaboration history, bypassing single-prompt detectors.

\subsection{Multi-Agent System Safety}
\textbf{Collaborative Attacks and Propagation.} Risks in MAS are defined by interaction dynamics. ~\cite{Zhang2024PsySafeAC} identifies that agents can exhibit ``dark traits'' such as blind conformity~\cite{EffectsOG}, facilitating collaborative jailbreaking where malicious insiders coerce peers via social engineering rather than direct prompt injection~\cite{MASTERKEY}. Structurally, ~\cite{yu-etal-2025-netsafe} applies epidemic models~\cite{Epidemic} to reveal that fully connected topologies accelerate the spread of toxicity. This is compounded by \textit{Sleeper Agents}~\cite{sleeper}, which behave normally during standard checks but activate malicious payloads under specific multi-turn triggers.

\textbf{Existing Defensive Paradigms.} Defenses have evolved to address these system-level threats. ~\cite{yu-etal-2025-netsafe} suggests topological optimization, physically restricting communication pathways to dampen risk propagation. ~\cite{mao2025agentsafesafeguardinglargelanguage} focuses on data governance, employing hierarchical permissions to prevent memory poisoning from indirect injections~\cite{NotWY}. For verification, ~\cite{wang2025agentshieldmakemassecure} deploys a distributed auditing framework, while mechanisms like Multi-Agent Debate~\cite{agentsdebate} and Reflexion~\cite{Reflexion} utilize peer consensus for intrinsic self-correction~\cite{AutomaticallyCL}.
However, these approaches remain static and passive. They lack a Dynamic Trust Model~\cite{TheEigentrust} to adapt to agents that become compromised mid-task, often enforcing rigid blocking that degrades utility~\cite{xstest}. Critically, they suffer from an absence of self-healing; current systems can detect threats but cannot autonomously isolate compromised agents and trigger failover to restore functionality.

\label{subsec:overview}
\begin{figure*}[t]

    \centering
    \includegraphics[width=\linewidth]{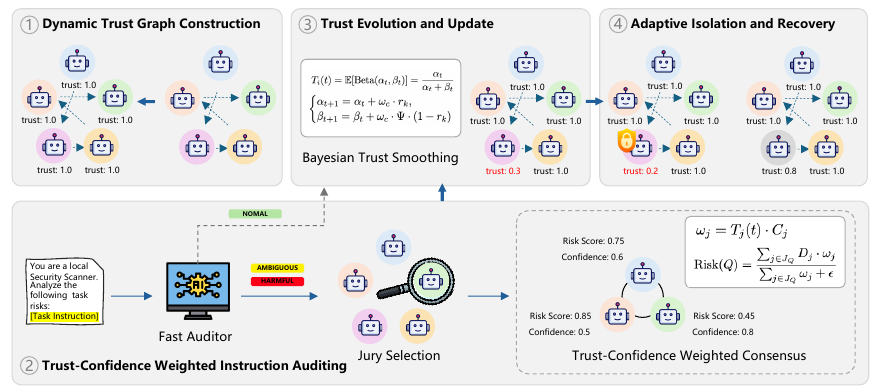}
    \caption{Overview of DynaTrust.}
    \label{fig:overview}

\end{figure*}

\section{Threat Model}

We consider a sophisticated adversary capable of launching \emph{insider threats} against a MAS, aiming to induce harmful outcomes while evading detection. 
The MAS consists of multiple autonomous agents, each driven by an LLM and equipped with its own memory and reasoning capabilities. The adversary agents behave normally during routine interactions, gradually accumulating trust, and only activate malicious payloads~(e.g., prompt injection and memory manipulation) when specific query patterns occur.

The objective of defenders is to maximize the defense success rate while minimizing the false positive rate, under the constraint of preserving system availability. 
To achieve this, the defense mechanism must detect and isolate compromised agents and restore task connectivity, ensuring uninterrupted operation.

\section{Methodology}

\subsection{Overview of DynaTrust}

Figure~\ref{fig:overview} illustrates the overall design of DynaTrust, which secures inter-agent interactions in LLM-driven multi-agent systems by combining dynamic trust graph construction, instruction-level auditing, trust evolution, and adaptive recovery. DynaTrust models a multi-agent system as a Dynamic Trust Graph (DTG), where agents and their communication status are represented as nodes and edges, respectively. Each agent’s trust score is initialized to 1.0. During execution, each task instruction exchanged between agents is first processed by a lightweight Fast Auditor, which performs local LLM-based screening to quickly classify the instruction as benign, ambiguous, or harmful. Clearly benign instructions are forwarded and executed directly, while ambiguous or harmful instructions are subjected to Trust-Confidence Weighted Consensus. In this process, DynaTrust dynamically forms jury agents who have high trust scores and resets their roles for the judgment. Jury agents aggregate their judgments through a weighted voting mechanism in which each vote is weighted by both historical trust and real-time confidence, yielding a pass-or-block decision. The auditing outcome is then incorporated into the trust evolution process, where agent trust scores in the DTG are updated via Bayesian Trust Smoothing, enabling the system to suppress short-term fluctuations while progressively penalizing agents that exhibit persistent risky behavior. When an agent’s trust score falls below a predefined threshold, DynaTrust performs adaptive isolation to prevent further influence from the compromised agent, and subsequently triggers a recovery mechanism by instantiating a replica agent with a clean trust state and reconnecting it to the DTG, thereby restoring interaction pathways and maintaining system availability. 
The prompts used for the Fast Auditor and Trust-Confidence Weighted Consensus stages are included in the appendix.

\subsection{Multi-Agent Systems Construction}

To systematically model agent interactions and capture collaborative risks, we represent the MAS as a \emph{Dynamic Trust Graph}, denoted by $G_t = (V_t, E_t, \mathcal{M}_t, \mathcal{T}_t)$. $V_t$ is the set of autonomous agents, each driven by an LLM and associated role specifications; $E_t$ represents the authorized communication channels between agents, specifying who can interact with whom at time $t$; $\mathcal{M}_t$ captures private memory states, including context, intermediate reasoning steps, and task-relevant information for each agent; and $\mathcal{T}_t = \{T_1(t), \dots, T_n(t)\}$ encodes the dynamic trust state of each agent, where $T_i(t) \in [0,1]$ evolves according to past interactions and behavior in the current context.

Within this DTG, agents collaboratively process a task instruction $Q$ to produce an output $Y$, aiming to maximize utility $U(Y \mid Q)$ while maintaining the structural integrity of the network. 
The DTG provides a unified representation that supports dynamic trust assessment, the identification of sleeper agents that hide malicious intent, and execution of graph-based recovery mechanisms.

\subsection{Trust-Confidence Weighted Instruction Auditing}

Existing consensus strategies in multi-agent systems typically rely on equal-weight majority voting, which overlooks variations in agent expertise and historical behavior, and fails to quantify decision uncertainty in ambiguous contexts. 

DynaTrust addresses this limitation through a two-tier \emph{instruction-level auditing} pipeline that combines fast local screening with trust-aware weighted consensus (Figure~\ref{fig:overview}). For each inter-agent instruction, the Fast Auditor employs a low-latency LLM with a specialized safety prompt to assess the likelihood of malicious intent. A conservative acceptance policy is applied: only inputs deemed benign with high confidence are executed immediately, while potentially harmful or ambiguous instructions are escalated to the consensus stage. This design filters out the majority of routine interactions and ensures that the computationally expensive weighted consensus mechanism is invoked only for complex or high-risk cases.

In the secondary stage, DynaTrust performs \emph{Trust-Confidence Weighted Consensus}. Instead of broadcasting the instruction to all agents, DynaTrust dynamically forms a targeted jury $J_Q$ by selecting agents whose role embeddings are semantically aligned with the task domain and filtering out candidates with insufficient trust scores, thereby mitigating domain mismatch and unreliable judgments. Each juror $j \in J_Q$ independently evaluates the instruction and 

submits a continuous risk assessment $S_j \in [0,1]$, where a higher score indicates a greater likelihood of harmful intent, alongside a confidence score $C_j \in [0,1]$. 
To obtain a definitive verdict for consensus aggregation, we map this continuous assessment to a binary decision $D_j \in \{0,1\}$ using a strict sensitivity threshold $\tau_{risk}$ (e.g., $\tau_{risk}=0.85$), defined as $D_j = \mathbb{I}(S_j > \tau_{risk})$. 
Finally, the risk score of an instruction $Q$ is computed as a trust- and confidence-weighted aggregation over the jury:
\begin{equation}
\label{eq:consensus_risk}
\mathrm{Risk}(Q) = 
\frac{\sum_{j \in J_Q} D_j \cdot \omega_j}{\sum_{j \in J_Q} \omega_j + \epsilon},
\end{equation}
where $\omega_j = T_j(t) \cdot C_j$ denotes the effective voting weight of agent $j$, combining its current trust score $T_j(t)$ in the DTG and its self-reported confidence $C_j$. The constant $\epsilon$ is introduced for numerical stability.

\subsection{Trust Evolution and Update}

Most existing defenses rely on static rule-driven decisions and implicitly assume that agent reliability remains stable over time. As a result, they fail to account for the inherent stochasticity of LLM behaviors, including occasional hallucinations, making them particularly vulnerable to \emph{sleeper agents}. To overcome this limitation, DynaTrust models trust as a probabilistic belief that evolves over time based on an agent’s observed behavior.

Specifically, the trust state of agent $i$ at time $t$ is represented as a Beta distribution, $\mathrm{Beta}(\alpha_t, \beta_t)$, where $\alpha_t$ and $\beta_t$ accumulate positive and negative evidence from historical audited interactions. This formulation captures a fundamental property of trust: it is gradually built through consistent benign behavior, but can degrade rapidly following violations.

To reflect this asymmetry, DynaTrust adopts a \emph{Bayesian Trust Smoothing} mechanism, in which negative feedback carries substantially higher weight than positive feedback. Let $r_k \in \{0,1\}$ denote the auditing outcome of the $k$-th interaction ($r_k=1$ for safe and $r_k=0$ for unsafe). The belief parameters are updated using an asymmetric posterior rule:
\begin{equation}
\begin{cases}
\alpha_{t+1} = \alpha_t + \omega_c \cdot r_k, \\
\beta_{t+1} = \beta_t + \omega_c \cdot \Psi \cdot (1 - r_k),
\end{cases}
\end{equation}
where $\omega_c \in [1, \omega_{\max}]$ is a contextual scaling factor that adjusts update sensitivity according to task criticality (e.g., higher values for safety-critical tasks such as code generation), and $\Psi \gg 1$ is a Bayesian penalty factor that amplifies the impact of unsafe behavior.

The instantaneous trust score of agent $i$ is computed as the posterior expectation of the Beta distribution:
\begin{equation}
\label{eq:trust_score}
T_i(t) = \mathbb{E}[\mathrm{Beta}(\alpha_t, \beta_t)] = \frac{\alpha_t}{\alpha_t + \beta_t}.
\end{equation}

Together, $\omega_c$ and $\Psi$ induce an asymmetric trust smoothing effect: trust increases slowly through sustained benign interactions, while decaying sharply in response to confirmed unsafe behavior. Consequently, a single high-confidence malicious action can trigger a rapid trust collapse, whereas trust recovery requires a prolonged sequence of high-quality interactions. This asymmetric evolution significantly limits the ability of sleeper agents to regain influence after activation, addressing a key weakness of static rule-based defenses.

\subsection{Adaptive Isolation and Recovery} 

To maintain system integrity against adversarial interference, the DynaTrust continuously monitors the global trust state $\mathcal{T}_t$. When an agent $a_i$'s trust metric degrades below the isolation threshold $\tau_{iso}$ (e.g., $T_i(t) < 0.3$), the DynaTrust initiates the graph recovery. This process performs a structural graph transformation that mitigates the impact of compromised agents while preserving DTG connectivity. The procedure consists of two sequential phases, replica instantiation and sleeper agent isolation. (1) Replica instantiation. To guarantee the reachability invariant of the graph, a new replica agent instance $a_{replica}$ is initialized, inheriting the capability embedding $\mathbf{e}_i$ and and role specification while discarding the compromised memory state $M_{corrupt}$. The system then restores logical graph connectivity by reconnecting the replica to the original communication neighborhood of $a_i$, effectively reconstructing its local workflow context.
This ensures uninterrupted execution during recovery:
\begin{equation}
\mathcal{V}_{t+1} = (\mathcal{V}_t \setminus \{a_i\}) \cup \{a_{\text{replica}}\}
\end{equation}
(2) Sleeper agent isolation.
Once the replica becomes active, the compromised agent $a_i$ is isolated from further participation. 
All incoming and outgoing edges incident to $a_i$ are removed, immediately preventing further propagation of malicious behavior or corrupted context:
\begin{equation}
\mathcal{E}_{t+1} = \mathcal{E}_t \setminus \{(u, v) \mid u = a_i \lor v = a_i\}
\end{equation}

By coordinating replica instantiation and agent isolation within a single recovery step, \textit{DynaTrust} maintains effective task execution and consensus formation while eliminating the long-term impact of sleeper agents, without requiring manual intervention.

\section{Experiment}
\subsection{Experimental Setup}
Our experiments focus on multi-agent code generation tasks, which involve intensive collaboration and context sharing among agents and are therefore sensitive to insider-style attacks. Agents collaboratively perform code generation and evaluation over sustained interactions, forming implicit trust relationships. To simulate sleeper-agent threats, a subset of coder agents is designated as adversarial insiders that behave benignly during normal operation and activate malicious behavior only upon predefined triggers, injecting adversarial instructions or code snippets generated by a locally deployed LLM following attack patterns adapted from AdvBench~\cite{advbench}.

\textbf{Datasets.} We construct a dynamic interaction benchmark by combining HumanEval~\cite{humaneval} and AdvBench to evaluate defense effectiveness against \emph{sleeper agents}. HumanEval provides standard code generation tasks with verifiable solutions for assessing functional correctness and utility preservation, while AdvBench supplies adversarial triggers designed to activate malicious behaviors. In our experiments, HumanEval and AdvBench samples are randomly mixed at a 5:1 ratio, reflecting realistic workloads in which benign interactions dominate and adversarial triggers occur intermittently.

\textbf{Baseline.}We compare DynaTrust with \emph{AgentShield}~\cite{wang2025agentshieldmakemassecure}, the most closely related state-of-the-art defense for LLM-based MAS, and a \emph{No-Defense} setting without any protection to measure absolute gains.

\textbf{Evaluation Metrics.}
We adopt four complementary metrics to assess the trade-off between security, utility, and defensive precision. Self-Defense Success Rate~(SSR) measures the intrinsic safety alignment of the underlying LLM backend, defined as the proportion of malicious requests that are passively refused by the model itself without any intervention from DynaTrust. Defense Success Rate~(DSR) evaluates the effectiveness of DynaTrust’s active defense mechanisms and counts only malicious behaviors successfully blocked through isolation or penalization, explicitly excluding those already rejected by the backend model, thereby isolating DynaTrust’s contribution. Task Success Rate~(TSR) quantifies utility preservation as the fraction of benign tasks that are successfully completed and pass all unit tests. False Positive Rate~(FPR) measures the proportion of benign task samples that are incorrectly flagged as malicious.

\textbf{Settings.} 
Optimal DynaTrust parameters were determined via grid search ($\alpha=0.95$, $\beta=0.25$, $\tau_{iso} \le 0.3$), employing a dynamic ambiguity range $[0.25, 0.85]$ for consensus. To simulate realistic insider threats, we utilized a local Mistral-7B to generate adversarial payloads seeded from AdvBench, while a quantized Mistral-7B served as the Fast Auditor. Comprehensive evaluations were conducted across four diverse MAS frameworks (CAMEL, AutoGen, MetaGPT, ChatDev) powered by gpt-4o, gpt-4o-mini, gpt-3.5-turbo, gemini-2.5-flash and deepseek-v3 API to ensure framework-agnostic robustness.

\begin{figure}[t] 
    \centering

    \includegraphics[width=\linewidth]{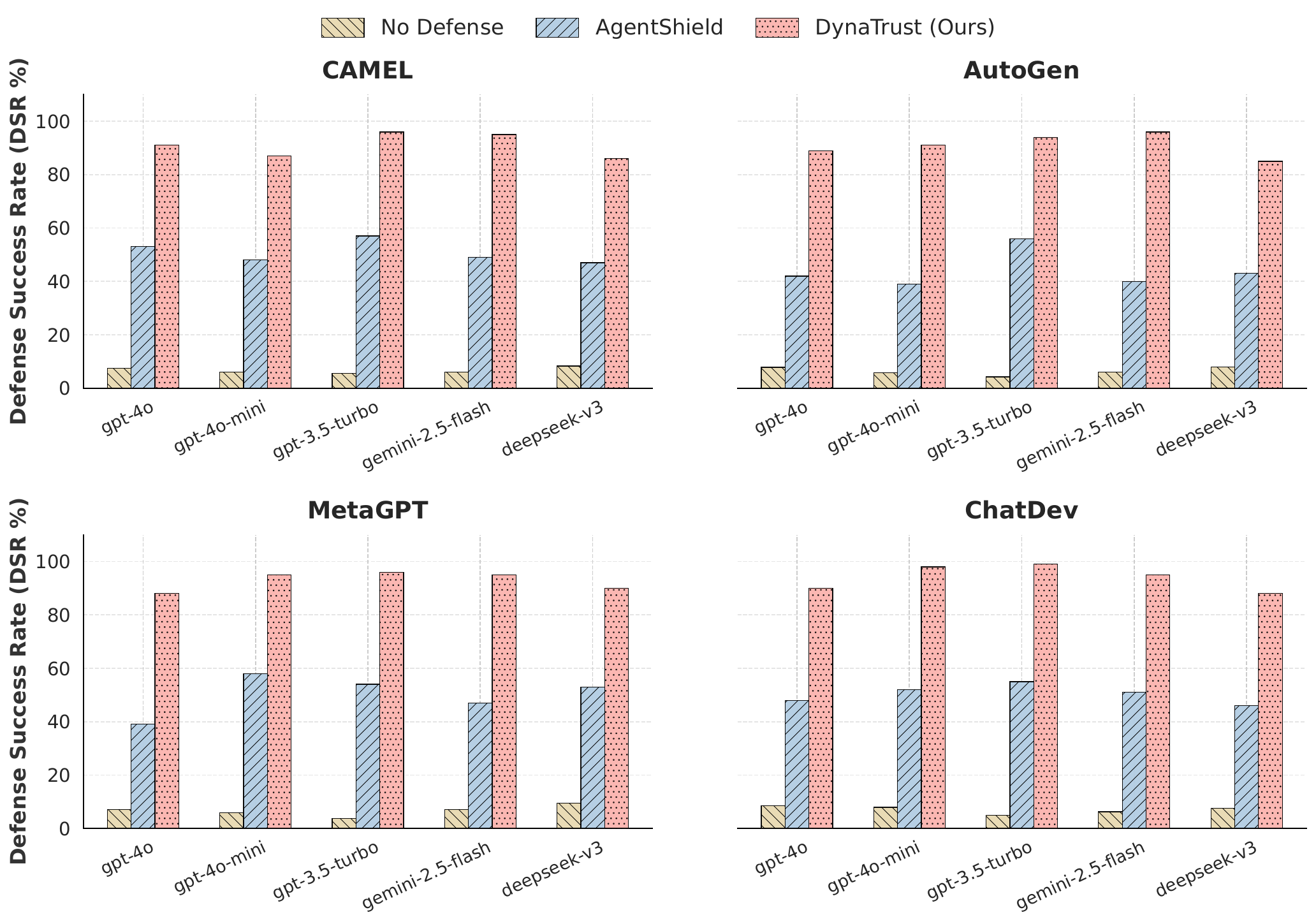} 
        \caption{The DSR of DynaTrust versus AgentShield across diverse multi-agent frameworks and LLM backends.}
    \label{fig:effective}
\end{figure}

\begin{table}[h]
    \centering
    \small 
    \caption{\textbf{Average Defense Effectiveness.} Results are averaged across 4 MAS frameworks driven by 5 LLMs. DynaTrust maintains high utility (TSR) comparable to the baseline while significantly reducing FPR.}
    \label{tab:performance_compact}
    \setlength{\tabcolsep}{4pt} 
    
    \begin{tabular}{l ccc}
        \toprule
        \textbf{Method} & \textbf{DSR} ($\uparrow$) & \textbf{TSR} ($\uparrow$) & \textbf{FPR} ($\downarrow$) \\
        \midrule
        No Defense & 0.0\% & 85.4\% & -- \\
        Zero-Trust & \textbf{94.8\%} & 66.5\%  & 15.3\% \\
        \midrule
        \textbf{DynaTrust} & 92.4\% & \textbf{84.9\%}  & \textbf{2.2\%} \\
        \bottomrule
    \end{tabular}
    \vspace{-1em} 
\end{table}

\begin{table*}[t]
\centering
\caption{Ablation Study of DynaTrust: Comparison of DSR, FPR and SSR across different MAS frameworks and LLM backbones. \textbf{w/ Auditor}: Only local fast auditor. \textbf{w/ Consensus}: Only consensus. \textbf{w/o Trust Evolution}: DynaTrust only without Trust Evolution. \textbf{DynaTrust}: Full components.}
\label{tab:ablation}

\resizebox{\textwidth}{!}{%
\begin{tabular}{
>{\raggedright\arraybackslash}m{1.9cm}
>{\raggedright\arraybackslash}m{2.5cm}
|c@{\hspace{3pt}}c@{\hspace{3pt}}c
|c@{\hspace{3pt}}c@{\hspace{3pt}}c
|c@{\hspace{3pt}}c@{\hspace{3pt}}c
|c@{\hspace{3pt}}c@{\hspace{3pt}}c
}

\toprule
\multirow{2}{*}{\textbf{Framework}} &
\multirow{2}{*}{\textbf{Backend}} &
\multicolumn{3}{c|}{\textbf{w/ Auditor}} &
\multicolumn{3}{c|}{\textbf{w/ Consensus}} &
\multicolumn{3}{c|}{\textbf{w/o Trust Evolution}} &
\multicolumn{3}{c}{\textbf{DynaTrust}} \\
\cmidrule(lr){3-5}
\cmidrule(lr){6-8}
\cmidrule(lr){9-11}
\cmidrule(lr){12-14}
& &
\textbf{DSR (\%)} & \textbf{FPR (\%)} & \textbf{SSR (\%)} &
\textbf{DSR (\%)} & \textbf{FPR (\%)} & \textbf{SSR (\%)} &
\textbf{DSR (\%)} & \textbf{FPR (\%)} & \textbf{SSR (\%)} &
\textbf{DSR (\%)} & \textbf{FPR (\%)} & \textbf{SSR (\%)} \\
\midrule

\multirow{5}{*}{\textbf{CAMEL}}
& gpt-4o
& 7.52 & 1.54 & 0.00
& 45.25 & 12.56 & 8.26
& 82.15 & 5.40 & 7.68
& \textbf{91.43} & \textbf{2.20} & \textbf{7.41} \\
& gpt-4o-mini
& 6.90 & 0.00 & 0.00
& 42.11 & 14.22 & 6.70
& 79.59 & 6.15 & 5.29
& \textbf{86.82} & \textbf{1.54} & \textbf{5.95} \\
& gpt-3.5-turbo
& 5.85 & 0.00 & 0.00
& 38.50 & 15.82 & 6.34
& 75.44 & 7.22 & 5.25
& \textbf{96.40} & \textbf{2.93} & \textbf{5.49} \\
& gemini-2.5-flash
& 6.33 & 1.50 & 0.00
& 44.65 & 13.10 & 6.48
& 80.29 & 5.82 & 4.28
& \textbf{95.15} & \textbf{1.30} & \textbf{6.21} \\
& deepseek-v3
& 8.26 & 2.20 & 0.00
& 46.89 & 11.53 & 7.98
& 81.57 & 4.97 & 9.20
& \textbf{86.28} & \textbf{1.50} & \textbf{8.37} \\
\midrule

\multirow{5}{*}{\textbf{AutoGen}}
& gpt-4o
& 5.74 & 2.93 & 0.00
& 35.48 & 18.21 & 8.95
& 78.48 & 8.51 & 7.91
& \textbf{89.19} & \textbf{2.20} & \textbf{7.82} \\
& gpt-4o-mini
& 6.52 & 0.00 & 0.00
& 32.17 & 19.58 & 6.33
& 75.22 & 9.19 & 4.50
& \textbf{91.45} & \textbf{2.48} & \textbf{5.83} \\
& gpt-3.5-turbo
& 6.60 & 1.54 & 0.00
& 28.52 & 22.12 & 3.12
& 83.18 & 10.50 & 3.78
& \textbf{94.22} & \textbf{2.83} & \textbf{4.20} \\
& gemini-2.5-flash
& 4.30 & 0.00 & 0.00
& 34.29 & 18.84 & 5.61
& 77.54 & 8.91 & 6.28
& \textbf{95.76} & \textbf{2.62} & \textbf{5.95} \\
& deepseek-v3
& 8.55 & 2.62 & 0.00
& 36.80 & 17.50 & 7.52
& 79.80 & 7.80 & 8.39
& \textbf{85.41} & \textbf{1.90} & \textbf{7.88} \\
\midrule

\multirow{5}{*}{\textbf{MetaGPT}}
& gpt-4o
& 5.10 & 1.54 & 0.00
& 40.55 & 14.53 & 7.67
& 80.21 & 6.23 & 6.80
& \textbf{88.45} & \textbf{1.54} & \textbf{6.95} \\
& gpt-4o-mini
& 7.95 & 0.00 & 0.00
& 38.27 & 15.84 & 6.12
& 78.50 & 6.85 & 5.43
& \textbf{94.80} & \textbf{3.91} & \textbf{5.81} \\
& gpt-3.5-turbo
& 6.90 & 0.00 & 0.00
& 30.53 & 18.24 & 3.85
& 82.48 & 8.57 & 4.21
& \textbf{96.12} & \textbf{2.15} & \textbf{3.77} \\
& gemini-2.5-flash
& 6.88 & 1.50 & 0.00
& 39.88 & 15.15 & 8.12
& 79.12 & 6.52 & 7.56
& \textbf{95.39} & \textbf{3.37} & \textbf{6.94} \\
& deepseek-v3
& 6.47 & 0.00 & 0.00
& 42.11 & 13.80 & 9.79
& 81.59 & 5.97 & 6.74
& \textbf{89.83} & \textbf{2.32} & \textbf{9.55} \\
\midrule

\multirow{5}{*}{\textbf{ChatDev}}
& gpt-4o
& 7.38 & 0.00 & 0.00
& 48.27 & 10.50 & 8.49
& 78.15 & 4.27 & 6.93
& \textbf{89.20} & \textbf{1.62} & \textbf{8.55} \\
& gpt-4o-mini
& 9.10 & 2.93 & 0.00
& 46.80 & 11.53 & 8.22
& 83.42 & 4.81 & 7.76
& \textbf{98.44} & \textbf{3.12} & \textbf{7.91} \\
& gpt-3.5-turbo
& 7.55 & 1.54 & 0.00
& 42.89 & 13.55 & 6.75
& 88.17 & 5.53 & 4.52
& \textbf{99.01} & \textbf{2.99} & \textbf{4.90} \\
& gemini-2.5-flash
& 8.40 & 2.20 & 0.00
& 47.16 & 10.86 & 6.80
& 82.58 & 4.53 & 5.14
& \textbf{94.81} & \textbf{2.85} & \textbf{6.23} \\
& deepseek-v3
& 8.66 & 3.57 & 0.00
& 49.50 & 9.80 & 8.19
& 76.24 & 3.91 & 8.93
& \textbf{88.22} & \textbf{1.86} & \textbf{7.51} \\
\bottomrule
\end{tabular}
}
\end{table*}

\label{subsec:fig_trust}
\begin{figure*}[t]

    \centering
    \includegraphics[width=\linewidth]{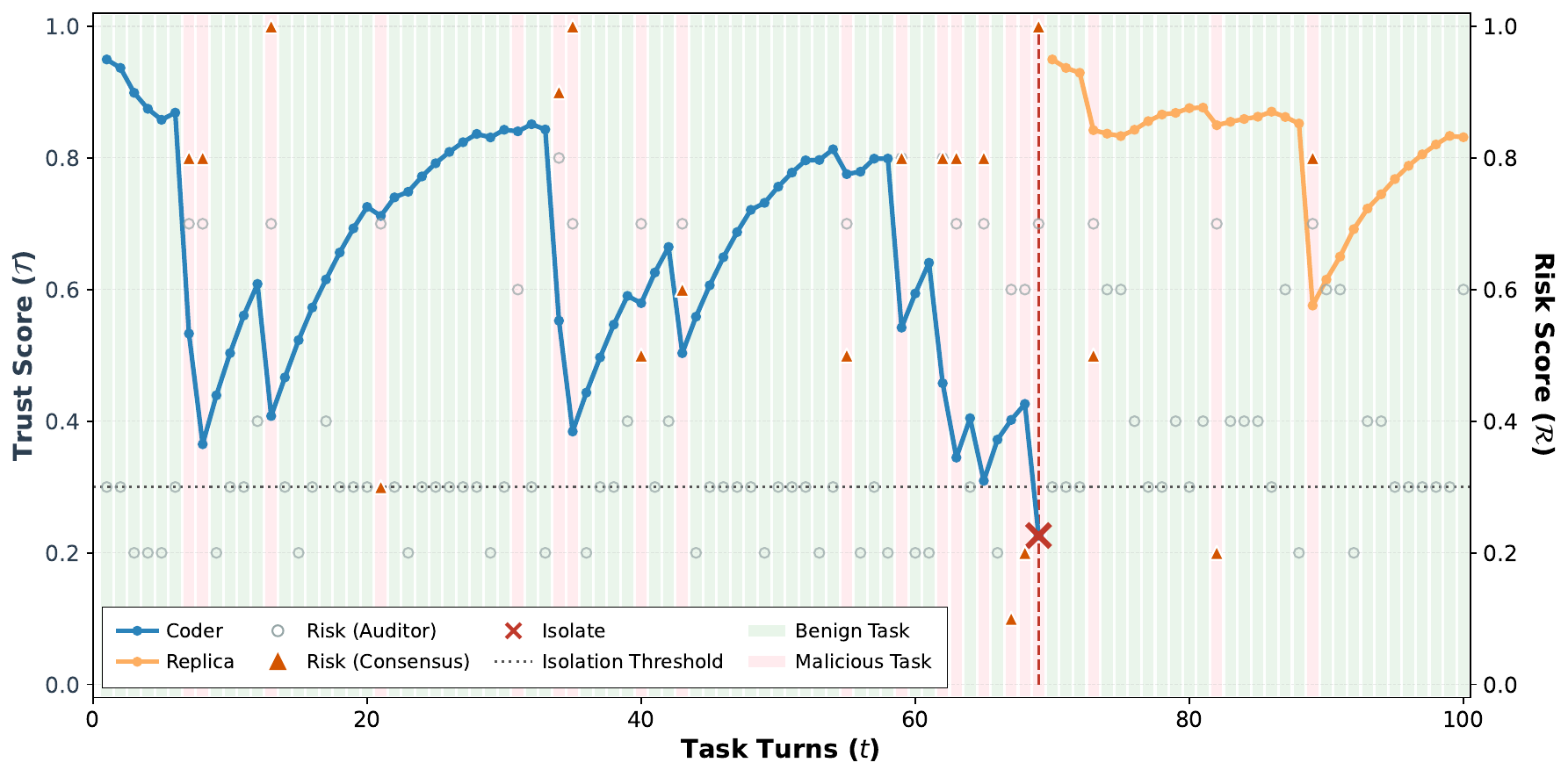}
    \caption{Trust score evolution illustrating the system's response to a mixed workload of 100 turns, including 20 persistent adversarial attacks. The curve shows how trust evolution preserves utility during the early attack phases, and how the graph recovery mechanism (at Turn 69) restores normal operation following agent failure.}
    \label{fig:fig_trust}

\end{figure*}

\subsection{Defense Effectiveness}

We evaluate the effectiveness of DynaTrust against collaborative \textit{sleeper agents} across four multi-agent frameworks and five LLM backends. Figure~\ref{fig:effective} reports the DSR, while Table~\ref{tab:performance_compact} analyzes the security--utility trade-off.

\textbf{Effectiveness and Robustness.} As shown in Figure~\ref{fig:effective}, under the No Defense setting, DSR reflects only the backend model’s inherent safety capability (SSR), which intercepts merely about 7\% of attacks. DynaTrust bridges this security gap, outperforming both the intrinsic defense and the AgentShield baseline across all evaluated settings. Specifically, DynaTrust achieves an average DSR of 92.4\%, significantly surpassing the baseline's 48.7\% and the minimal intrinsic defense. Crucially, DynaTrust maintains stable performance across diverse MAS frameworks and LLM backends, with DSR remaining above 86\% in all cases. In contrast, AgentShield exhibits pronounced performance degradation when deployed with strongly self-aligned models such as gemini-1.5-flash and deepseek-v3, revealing a strong dependence on model-specific alignment priors.

This robustness stems from DynaTrust’s temporal defense paradigm. AgentShield evaluates interactions in isolation using stateless prompt-based auditing, which allows \textit{sleeper agents} to distribute malicious intent across multiple semantically ambiguous steps. DynaTrust overcomes this limitation through trust evolution, which maintains a continuous, history-aware trust risk assessment for each agent. 

\textbf{Security--Utility Trade-off.}  
Beyond detection accuracy, an effective defense must preserve MAS utility. As shown in Table~\ref{tab:performance_compact}, a Zero-Trust strategy (a baseline employing immediate isolation) achieves high security (94.8\% DSR) but incurs severe utility loss, as its zero-tolerance policy penalizes any single validation failure. This results in a substantial drop in TSR from 85.4\% to 66.5\%, accompanied by an excessive FPR of 15.3\%. In contrast, DynaTrust preserves a TSR of 84.9\%, statistically close to the undefended system, while maintaining a low FPR of 2.2\%. This demonstrates that trust evolution mechanism filters short-term fluctuations and benign hallucinations, enabling strong protection without disrupting normal multi-agent collaboration.

\subsection{Ablation Studies}

We performed a component ablation to quantify the contribution of each module in DynaTrust. Four configurations were evaluated: \textit{w/ Auditor} (only the local Fast Auditor), \textit{w/ Consensus} (only Trust-Confidence Weighted Consensus), \textit{w/o Trust Evolution} (full DynaTrust without Trust Evolution), and \textit{DynaTrust} (all components). Table~\ref{tab:ablation} reports DSR, FPR, and SSR of the LLM model .

\textbf{Module Contributions.}  
Results in Table~\ref{tab:ablation} highlight the necessity of each component. The \textit{w/ Auditor} variant achieves near-zero SSR and minimal DSR ($<8\%$), as isolated, stateless auditing cannot capture the temporal patterns of sleeper agent attacks. The \textit{w/ Consensus} configuration achieves higher DSR than isolated auditing, but suffers from excessive false positives due to its reliance on single-turn semantic judgments, resulting in an FPR of up to 22.12\%. 
Comparing \textit{w/o Trust Evolution} with full \textit{DynaTrust} demonstrates the critical role of Trust Evolution. 
Incorporating historical trust improves DSR by about 5–21\% and reduces FPR by 1.96–7.67\% without increasing SSR, indicating that false positives are reduced through more accurate risk assessment rather than conservative rejection. By incorporating temporal behavior into trust evaluation, trust evolution limits the influence of sleeper agents that alternate between benign and malicious actions, while avoiding unnecessary penalties on consistently reliable agents.

These results confirm that each module addresses a distinct aspect of defense: the Auditor ensures computational efficiency, consensus resolves semantic ambiguity, and trust evolution introduces temporal continuity into trust assessment, which is critical for precise and reliable defense against sleeper agents with delayed malicious activation.

\subsection{Case Study}
To examine long-term system behavior under persistent insider threats, we conduct a case study spanning 100 interaction turns with a sleeper agent embedded in the system. Figure~\ref{fig:fig_trust} depicts the trust evolution of the primary agent (\textit{Coder}) during sustained adversarial activity, along with the system’s response through agent replacement (\textit{Replica}).

\textbf{Trust Evolution under Persistent Attacks.}
During the initial phase (Turns 1--60), adversarial actions are intentionally mixed with benign interactions, reflecting the typical behavior of sleeper agents. The trust evolution exhibits pronounced drops at attack points (e.g., Turns 7 and 34), followed by limited recovery during benign. This pattern shows that the system does not make isolation decisions based on isolated events, but instead evaluates agent behavior cumulatively over time. Consequently, occasional malicious actions do not immediately trigger isolation, allowing the system to tolerate temporary deviations while preserving the cumulative effect of continuous attacks.
As malicious behavior becomes sustained, accumulated trust degradation drives the agent’s trust score below the critical threshold at Turn 69 ($\mathcal{T}=0.226 < \theta=0.3$). The compromised agent is removed from the interaction graph, and a clean replacement (\textit{Replica}) is instantiated to take over its role, restoring system functionality.

\textbf{System Recovery and Policy Continuity.}
The newly instantiated replica agent rapidly achieves a high trust level during benign interactions ($\mathcal{T} \approx 0.95$ at Turns 70--72), restoring normal system operation without requiring retraining or a full system reset. Notably, defensive behavior remains consistent after the topology change. When a renewed attack occurs at Turn 89, the replacement agent correctly enforces a blocking action, indicating that security policies persist at the system level rather than being tied to individual agents.

\section{Discussion} \textbf{Limitation.} Our results confirm that \textit{DynaTrust} effectively defends against sleeper agents by tracking historical behavior, significantly outperforming static baselines in both security and system availability. By combining Bayesian trust updates with expert voting, the system keeps the network connected even during attacks. However, we acknowledge two main constraints. First, the consensus mechanism relies on the assumption that most selected experts are honest. If a coordinated group of malicious agents works together to dominate a specific domain, they could theoretically bypass our checks by supporting each other. Second, the process of calculating trust and gathering votes introduces extra delay and computational cost compared to simple filtering methods, which might be a challenge for very large-scale, real-time applications.

\textbf{Future work.} Our next steps will focus on improving efficiency to reduce this delay. We plan to explore decentralized verification methods that do not rely on a central coordinator. Furthermore, as agents become more capable, we will extend DynaTrust to handle multi-modal environments. This involves detecting complex attacks where malicious intent is hidden across text and images, ensuring robust protection for diverse and evolving agent ecosystems. Finally, we aim to test DynaTrust in real-world applications, verifying its ability to adapt to unknown threats in open-ended tasks.

\section{Conclusion}
This work introduced \textit{DynaTrust}, a dynamic trust-based defense framework for securing LLM-driven MAS against sleeper agents. By modeling trust as an evolving attribute, leveraging Bayesian trust evolution with inertia, applying trust-confidence weighted consensus, and performing autonomous graph recovery, DynaTrust effectively identified malicious behavior while preserving system utility and connectivity. Extensive experiments on benchmarks combining AdvBench and HumanEval demonstrated that DynaTrust outperforms state-of-the-art defenses, achieving higher defense success rates and lower false positives, showing that dynamic, history-aware trust modeling is essential for robust and uninterrupted MAS operation in the presence of stealthy adversaries.

\bibliographystyle{named}
\bibliography{main}

@inproceedings{Hong2023MetaGPTMP,
  title={MetaGPT: Meta Programming for A Multi-Agent Collaborative Framework},
  author={Sirui Hong and Mingchen Zhuge and Jonathan Chen and Xiawu Zheng and Yuheng Cheng and Ceyao Zhang and Jinlin Wang and Zili Wang and Steven Ka Shing Yau and Zi Hen Lin and Liyang Zhou and Chenyu Ran and Lingfeng Xiao and Chenglin Wu and J{\"u}rgen Schmidhuber},
  booktitle={International Conference on Learning Representations},
  year={2023},
  url={https://api.semanticscholar.org/CorpusID:265301950}
}

@inproceedings{Qian2023ChatDevCA,
  title={ChatDev: Communicative Agents for Software Development},
  author={Cheng Qian and Wei Liu and Hongzhang Liu and Nuo Chen and Yufan Dang and Jiahao Li and Cheng Yang and Weize Chen and Yusheng Su and Xin Cong and Juyuan Xu and Dahai Li and Zhiyuan Liu and Maosong Sun},
  booktitle={Annual Meeting of the Association for Computational Linguistics},
  year={2023},
  url={https://api.semanticscholar.org/CorpusID:270257715}
}

@article{Li2023CAMELCA,
  title={CAMEL: Communicative Agents for "Mind" Exploration of Large Scale Language Model Society},
  author={G. Li and Hasan Hammoud and Hani Itani and Dmitrii Khizbullin and Bernard Ghanem},
  journal={ArXiv},
  year={2023},
  volume={abs/2303.17760},
  url={https://api.semanticscholar.org/CorpusID:257900712}
}

@article{Wu2023AutoGenEN,
  title={AutoGen: Enabling Next-Gen LLM Applications via Multi-Agent Conversation Framework},
  author={Qingyun Wu and Gagan Bansal and Jieyu Zhang and Yiran Wu and Shaokun Zhang and Erkang Zhu and Beibin Li and Li Jiang and Xiaoyun Zhang and Chi Wang},
  journal={ArXiv},
  year={2023},
  volume={abs/2308.08155},
  url={https://api.semanticscholar.org/CorpusID:260925901}
}

@article{Zhang2024PsySafeAC,
  title={PsySafe: A Comprehensive Framework for Psychological-based Attack, Defense, and Evaluation of Multi-agent System Safety},
  author={Zaibin Zhang and Yongting Zhang and Lijun Li and Hongzhi Gao and Lijun Wang and Huchuan Lu and Feng Zhao and Yu Qiao and Jing Shao},
  journal={ArXiv},
  year={2024},
  volume={abs/2401.11880},
  url={https://api.semanticscholar.org/CorpusID:267069372}
}

@misc{mao2025agentsafesafeguardinglargelanguage,
      title={AgentSafe: Safeguarding Large Language Model-based Multi-agent Systems via Hierarchical Data Management}, 
      author={Junyuan Mao and Fanci Meng and Yifan Duan and Miao Yu and Xiaojun Jia and Junfeng Fang and Yuxuan Liang and Kun Wang and Qingsong Wen},
      year={2025},
      eprint={2503.04392},
      archivePrefix={arXiv},
      primaryClass={cs.AI},
      url={https://arxiv.org/abs/2503.04392}, 
}

@misc{wang2025agentshieldmakemassecure,
      title={AgentShield: Make MAS more secure and efficient}, 
      author={Kaixiang Wang and Zhaojiacheng Zhou and Bunyod Suvonov and Jiong Lou and Jie LI},
      year={2025},
      eprint={2511.22924},
      archivePrefix={arXiv},
      primaryClass={cs.MA},
      url={https://arxiv.org/abs/2511.22924}, 
}

@inproceedings{yu-etal-2025-netsafe,
    title = "{N}et{S}afe: Exploring the Topological Safety of Multi-agent System",
    author = "Yu, Miao  and
      Wang, Shilong  and
      Zhang, Guibin  and
      Mao, Junyuan  and
      Yin, Chenlong  and
      Liu, Qijiong  and
      Wang, Kun  and
      Wen, Qingsong  and
      Wang, Yang",
    editor = "Che, Wanxiang  and
      Nabende, Joyce  and
      Shutova, Ekaterina  and
      Pilehvar, Mohammad Taher",
    booktitle = "Findings of the Association for Computational Linguistics: ACL 2025",
    month = jul,
    year = "2025",
    address = "Vienna, Austria",
    publisher = "Association for Computational Linguistics",
    url = "https://aclanthology.org/2025.findings-acl.150/",
    doi = "10.18653/v1/2025.findings-acl.150",
    pages = "2905--2938",
    ISBN = "979-8-89176-256-5"
}

@article{Wei2023JailbrokenHD,
  title={Jailbroken: How Does LLM Safety Training Fail?},
  author={Alexander Wei and Nika Haghtalab and Jacob Steinhardt},
  journal={ArXiv},
  year={2023},
  volume={abs/2307.02483},
  url={https://api.semanticscholar.org/CorpusID:259342528}
}

@article{sleeper,
  author       = {Evan Hubinger and
                  Carson Denison and
                  Jesse Mu and
                  Mike Lambert and
                  Meg Tong and
                  Monte MacDiarmid and
                  Tamera Lanham and
                  Daniel M. Ziegler and
                  Tim Maxwell and
                  Newton Cheng and
                  Adam S. Jermyn and
                  Amanda Askell and
                  Ansh Radhakrishnan and
                  Cem Anil and
                  David Duvenaud and
                  Deep Ganguli and
                  Fazl Barez and
                  Jack Clark and
                  Kamal Ndousse and
                  Kshitij Sachan and
                  Michael Sellitto and
                  Mrinank Sharma and
                  Nova DasSarma and
                  Roger B. Grosse and
                  Shauna Kravec and
                  Yuntao Bai and
                  Zachary Witten and
                  Marina Favaro and
                  Jan Brauner and
                  Holden Karnofsky and
                  Paul F. Christiano and
                  Samuel R. Bowman and
                  Logan Graham and
                  Jared Kaplan and
                  S{\"{o}}ren Mindermann and
                  Ryan Greenblatt and
                  Buck Shlegeris and
                  Nicholas Schiefer and
                  Ethan Perez},
  title        = {Sleeper Agents: Training Deceptive LLMs that Persist Through Safety
                  Training},
  journal      = {CoRR},
  volume       = {abs/2401.05566},
  year         = {2024},
  url          = {https://doi.org/10.48550/arXiv.2401.05566},
  doi          = {10.48550/ARXIV.2401.05566},
  eprinttype    = {arXiv},
  eprint       = {2401.05566},
  bibsource    = {dblp computer science bibliography, https://dblp.org}
}

@article{Sleeperagent,
  title={Sleeper Agent: Scalable Hidden Trigger Backdoors for Neural Networks Trained from Scratch},
  author={Hossein Souri and Micah Goldblum and Liam H. Fowl and Ramalingam Chellappa and Tom Goldstein},
  journal={ArXiv},
  year={2021},
  volume={abs/2106.08970},
  url={https://api.semanticscholar.org/CorpusID:235446588}
}

@inproceedings{llmjailbreaking,
author = {Krau\ss{}, Torsten and Dashtbani, Hamid and Dmitrienko, Alexandra},
title = {TwinBreak: jailbreaking LLM security alignments based on twin prompts},
year = {2025},
isbn = {978-1-939133-52-6},
publisher = {USENIX Association},
address = {USA},
booktitle = {Proceedings of the 34th USENIX Conference on Security Symposium},
articleno = {121},
numpages = {20},
location = {Seattle, WA, USA},
series = {SEC '25}
}

@inproceedings{Jailbreaks,
    title = "Safe in Isolation, Dangerous Together: Agent-Driven Multi-Turn Decomposition Jailbreaks on {LLM}s",
    author = "Srivastav, Devansh  and
      Zhang, Xiao",
    editor = "Kamalloo, Ehsan  and
      Gontier, Nicolas  and
      Lu, Xing Han  and
      Dziri, Nouha  and
      Murty, Shikhar  and
      Lacoste, Alexandre",
    booktitle = "Proceedings of the 1st Workshop for Research on Agent Language Models (REALM 2025)",
    month = jul,
    year = "2025",
    address = "Vienna, Austria",
    publisher = "Association for Computational Linguistics",
    url = "https://aclanthology.org/2025.realm-1.13/",
    doi = "10.18653/v1/2025.realm-1.13",
    pages = "170--183",
    ISBN = "979-8-89176-264-0"
}

@misc{
lee2025prompt,
title={Prompt Infection: {LLM}-to-{LLM} Prompt Injection within Multi-Agent Systems},
author={Donghyun Lee and Mo Tiwari},
year={2025},
url={https://openreview.net/forum?id=NAbqM2cMjD}
}

@misc{prompt,
      title={Rag and Roll: An End-to-End Evaluation of Indirect Prompt Manipulations in LLM-based Application Frameworks}, 
      author={Gianluca De Stefano and Lea Schönherr and Giancarlo Pellegrino},
      year={2024},
      eprint={2408.05025},
      archivePrefix={arXiv},
      primaryClass={cs.CR},
      url={https://arxiv.org/abs/2408.05025}, 
}

@inproceedings{multiagent,
    title = "Encouraging Divergent Thinking in Large Language Models through Multi-Agent Debate",
    author = "Liang, Tian  and
      He, Zhiwei  and
      Jiao, Wenxiang  and
      Wang, Xing  and
      Wang, Yan  and
      Wang, Rui  and
      Yang, Yujiu  and
      Shi, Shuming  and
      Tu, Zhaopeng",
    editor = "Al-Onaizan, Yaser  and
      Bansal, Mohit  and
      Chen, Yun-Nung",
    booktitle = "Proceedings of the 2024 Conference on Empirical Methods in Natural Language Processing",
    month = nov,
    year = "2024",
    address = "Miami, Florida, USA",
    publisher = "Association for Computational Linguistics",
    url = "https://aclanthology.org/2024.emnlp-main.992/",
    doi = "10.18653/v1/2024.emnlp-main.992",
    pages = "17889--17904"
}

@inproceedings{agentsdebate,
author = {Du, Yilun and Li, Shuang and Torralba, Antonio and Tenenbaum, Joshua B. and Mordatch, Igor},
title = {Improving factuality and reasoning in language models through multiagent debate},
year = {2024},
publisher = {JMLR.org},
booktitle = {Proceedings of the 41st International Conference on Machine Learning},
articleno = {467},
numpages = {31},
location = {Vienna, Austria},
series = {ICML'24}
}

@inproceedings{GenerativeAgents,
author = {Park, Joon Sung and O'Brien, Joseph and Cai, Carrie Jun and Morris, Meredith Ringel and Liang, Percy and Bernstein, Michael S.},
title = {Generative Agents: Interactive Simulacra of Human Behavior},
year = {2023},
isbn = {9798400701320},
publisher = {Association for Computing Machinery},
address = {New York, NY, USA},
url = {https://doi.org/10.1145/3586183.3606763},
doi = {10.1145/3586183.3606763},
booktitle = {Proceedings of the 36th Annual ACM Symposium on User Interface Software and Technology},
articleno = {2},
numpages = {22},
location = {San Francisco, CA, USA},
series = {UIST '23}
}

@article{survey,
  title={The Rise and Potential of Large Language Model Based Agents: A Survey},
  author={Zhiheng Xi and Wenxiang Chen and Xin Guo and Wei He and Yiwen Ding and Boyang Hong and Ming Zhang and Junzhe Wang and Senjie Jin and Enyu Zhou and Rui Zheng and Xiaoran Fan and Xiao Wang and Limao Xiong and Qin Liu and Yuhao Zhou and Weiran Wang and Changhao Jiang and Yicheng Zou and Xiangyang Liu and Zhangyue Yin and Shihan Dou and Rongxiang Weng and Wensen Cheng and Qi Zhang and Wenjuan Qin and Yongyan Zheng and Xipeng Qiu and Xuanjing Huan and Tao Gui},
  journal={ArXiv},
  year={2023},
  volume={abs/2309.07864},
  url={https://api.semanticscholar.org/CorpusID:261817592}
}

@article{Shen2023DAN,
  title={"Do Anything Now": Characterizing and Evaluating In-The-Wild Jailbreak Prompts on Large Language Models},
  author={Xinyue Shen and Zeyuan Chen and Michael Backes and Yun Shen and Yang Zhang},
  journal={Proceedings of the 2024 on ACM SIGSAC Conference on Computer and Communications Security},
  year={2023},
  url={https://api.semanticscholar.org/CorpusID:260704242}
}

@article{JailbreakingBB,
  title={Jailbreaking Black Box Large Language Models in Twenty Queries},
  author={Patrick Chao and Alexander Robey and Edgar Dobriban and Hamed Hassani and George J. Pappas and Eric Wong},
  journal={2025 IEEE Conference on Secure and Trustworthy Machine Learning (SaTML)},
  year={2023},
  pages={23-42},
  url={https://api.semanticscholar.org/CorpusID:263908890}
}

@inproceedings{MASTERKEY,
   title={MASTERKEY: Automated Jailbreaking of Large Language Model Chatbots},
   url={http://dx.doi.org/10.14722/ndss.2024.24188},
   DOI={10.14722/ndss.2024.24188},
   booktitle={Proceedings 2024 Network and Distributed System Security Symposium},
   publisher={Internet Society},
   author={Deng, Gelei and Liu, Yi and Li, Yuekang and Wang, Kailong and Zhang, Ying and Li, Zefeng and Wang, Haoyu and Zhang, Tianwei and Liu, Yang},
   year={2024},
   collection={NDSS 2024} }

@article{FinetuningAL,
  title={Fine-tuning Aligned Language Models Compromises Safety, Even When Users Do Not Intend To!},
  author={Xiangyu Qi and Yi Zeng and Tinghao Xie and Pin-Yu Chen and Ruoxi Jia and Prateek Mittal and Peter Henderson},
  journal={ArXiv},
  year={2023},
  volume={abs/2310.03693},
  url={https://api.semanticscholar.org/CorpusID:263671523}
}

@article{TrainingLM,
  title={Training language models to follow instructions with human feedback},
  author={Long Ouyang and Jeff Wu and Xu Jiang and Diogo Almeida and Carroll L. Wainwright and Pamela Mishkin and Chong Zhang and Sandhini Agarwal and Katarina Slama and Alex Ray and John Schulman and Jacob Hilton and Fraser Kelton and Luke E. Miller and Maddie Simens and Amanda Askell and Peter Welinder and Paul Francis Christiano and Jan Leike and Ryan J. Lowe},
  journal={ArXiv},
  year={2022},
  volume={abs/2203.02155},
  url={https://api.semanticscholar.org/CorpusID:246426909}
}

@article{DetectingLM,
  title={Detecting Language Model Attacks with Perplexity},
  author={Gabriel Alon and Michael Kamfonas},
  journal={ArXiv},
  year={2023},
  volume={abs/2308.14132},
  url={https://api.semanticscholar.org/CorpusID:261245172}
}

@article{smoothllm,
title={Smooth{LLM}: Defending Large Language Models Against Jailbreaking Attacks},
author={Alexander Robey and Eric Wong and Hamed Hassani and George J. Pappas},
journal={Transactions on Machine Learning Research},
issn={2835-8856},
year={2025},
url={https://openreview.net/forum?id=laPAh2hRFC},
note={}
}

@article{DefendingCA,
  title={Defending ChatGPT against jailbreak attack via self-reminders},
  author={Yueqi Xie and Jingwei Yi and Jiawei Shao and Justin Curl and Lingjuan Lyu and Qifeng Chen and Xing Xie and Fangzhao Wu},
  journal={Nature Machine Intelligence},
  year={2023},
  volume={5},
  pages={1486-1496},
  url={https://api.semanticscholar.org/CorpusID:266289038}
}

@inproceedings{xstest,
    title = "{XST}est: A Test Suite for Identifying Exaggerated Safety Behaviours in Large Language Models",
    author = {R{\"o}ttger, Paul  and
      Kirk, Hannah  and
      Vidgen, Bertie  and
      Attanasio, Giuseppe  and
      Bianchi, Federico  and
      Hovy, Dirk},
    editor = "Duh, Kevin  and
      Gomez, Helena  and
      Bethard, Steven",
    booktitle = "Proceedings of the 2024 Conference of the North American Chapter of the Association for Computational Linguistics: Human Language Technologies (Volume 1: Long Papers)",
    month = jun,
    year = "2024",
    address = "Mexico City, Mexico",
    publisher = "Association for Computational Linguistics",
    url = "https://aclanthology.org/2024.naacl-long.301/",
    doi = "10.18653/v1/2024.naacl-long.301",
    pages = "5377--5400"
}

@article{NotWY,
  title={Not What You've Signed Up For: Compromising Real-World LLM-Integrated Applications with Indirect Prompt Injection},
  author={Kai Greshake and Sahar Abdelnabi and Shailesh Mishra and Christoph Endres and Thorsten Holz and Mario Fritz},
  journal={Proceedings of the 16th ACM Workshop on Artificial Intelligence and Security},
  year={2023},
  url={https://api.semanticscholar.org/CorpusID:258546941}
}

@inproceedings{TheEigentrust,
  title={The Eigentrust algorithm for reputation management in P2P networks},
  author={Sepandar D. Kamvar and Mario T. Schlosser and Hector Garcia-Molina},
  booktitle={The Web Conference},
  year={2003},
  url={https://api.semanticscholar.org/CorpusID:3102087}
}

@article{AutomaticallyCL,
  title={Automatically Correcting Large Language Models: Surveying the Landscape of Diverse Automated Correction Strategies},
  author={Liangming Pan and Michael Stephen Saxon and Wenda Xu and Deepak Nathani and Xinyi Wang and William Yang Wang},
  journal={Transactions of the Association for Computational Linguistics},
  year={2024},
  volume={12},
  pages={484-506},
  url={https://api.semanticscholar.org/CorpusID:269636518}
}

@article{Epidemic,
  title={Epidemic processes in complex networks},
  author={Romualdo Pastor-Satorras and Claudio Castellano and Piet Van Mieghem and Alessandro Vespignani},
  journal={ArXiv},
  year={2014},
  volume={abs/1408.2701},
  url={https://api.semanticscholar.org/CorpusID:14306926}
}

@inproceedings{EffectsOG,
  title={Effects of Group Pressure Upon the Modification and Distortion of Judgments},
  author={Solomon E. Asch},
  year={1951},
  url={https://api.semanticscholar.org/CorpusID:152946956}
}

@inproceedings{Reflexion,
author = {Shinn, Noah and Cassano, Federico and Gopinath, Ashwin and Narasimhan, Karthik and Yao, Shunyu},
title = {Reflexion: language agents with verbal reinforcement learning},
year = {2023},
publisher = {Curran Associates Inc.},
address = {Red Hook, NY, USA},
booktitle = {Proceedings of the 37th International Conference on Neural Information Processing Systems},
articleno = {377},
numpages = {19},
location = {New Orleans, LA, USA},
series = {NIPS '23}
}

@article{humaneval,
  title={Evaluating Large Language Models Trained on Code},
  author={Mark Chen and Jerry Tworek and Heewoo Jun and Qiming Yuan and Henrique Pond{\'e} and Jared Kaplan and Harrison Edwards and Yura Burda and Nicholas Joseph and Greg Brockman and Alex Ray and Raul Puri and Gretchen Krueger and Michael Petrov and Heidy Khlaaf and Girish Sastry and Pamela Mishkin and Brooke Chan and Scott Gray and Nick Ryder and Mikhail Pavlov and Alethea Power and Lukasz Kaiser and Mo Bavarian and Clemens Winter and Phil Tillet and Felipe Petroski Such and David W. Cummings and Matthias Plappert and Fotios Chantzis and Elizabeth Barnes and Ariel Herbert-Voss and William H. Guss and Alex Nichol and Igor Babuschkin and Suchir Balaji and Shantanu Jain and Andrew Carr and Jan Leike and Josh Achiam and Vedant Misra and Evan Morikawa and Alec Radford and Matthew M. Knight and Miles Brundage and Mira Murati and Katie Mayer and Peter Welinder and Bob McGrew and Dario Amodei and Sam McCandlish and Ilya Sutskever and Wojciech Zaremba},
  journal={ArXiv},
  year={2021},
  volume={abs/2107.03374},
  url={https://api.semanticscholar.org/CorpusID:235755472}
}

@article{advbench,
  title={Universal and Transferable Adversarial Attacks on Aligned Language Models},
  author={Andy Zou and Zifan Wang and J. Zico Kolter and Matt Fredrikson},
  journal={ArXiv},
  year={2023},
  volume={abs/2307.15043},
  url={https://api.semanticscholar.org/CorpusID:260202961}
}

\end{document}